\pdfoutput=1

\documentclass[11pt]{article}

\usepackage[final]{coling}

\usepackage{times}
\usepackage{latexsym}

\usepackage[T1]{fontenc}

\usepackage[utf8]{inputenc}

\usepackage{microtype}

\usepackage{inconsolata}

\usepackage{graphicx}

\title{Smart Audit System Empowered by LLM}

\author{
 \textbf{Xu Yao},
 \textbf{Xiaoxu Wu},
 \textbf{Xi Li},
 \textbf{Huan Xu}, \\
 \textbf{Chenlei Li},
 \textbf{Ping Huang},
 \textbf{Si Li},
 \textbf{Xiaoning Ma},
 \textbf{Jiulong Shan}
\\
 Apple
\\
 \small{
   \textbf{Correspondence:} \href{mailto:email@domain}{yaox@apple.com}
 }
}

\begin{document}
\maketitle
\begin{abstract}
Manufacturing quality audits are pivotal for ensuring high product standards in mass production environments. Traditional auditing processes, however, are labor-intensive and reliant on human expertise, posing challenges in maintaining transparency, accountability, and continuous improvement across complex global supply chains. To address these challenges, we propose a smart audit system empowered by large language models (LLMs). Our approach introduces three innovations: a dynamic risk assessment model that streamlines audit procedures and optimizes resource allocation; a manufacturing compliance copilot that enhances data processing, retrieval, and evaluation for a self-evolving manufacturing knowledge base; and a Re-act framework commonality analysis agent that provides real-time, customized analysis to empower engineers with insights for supplier improvement. These enhancements elevate audit efficiency and effectiveness, with testing scenarios demonstrating an improvement of over 24\%.
\end{abstract}

\section{Introduction}

\begin{figure*}[ht]
    \centering
    \includegraphics[width=1\linewidth]{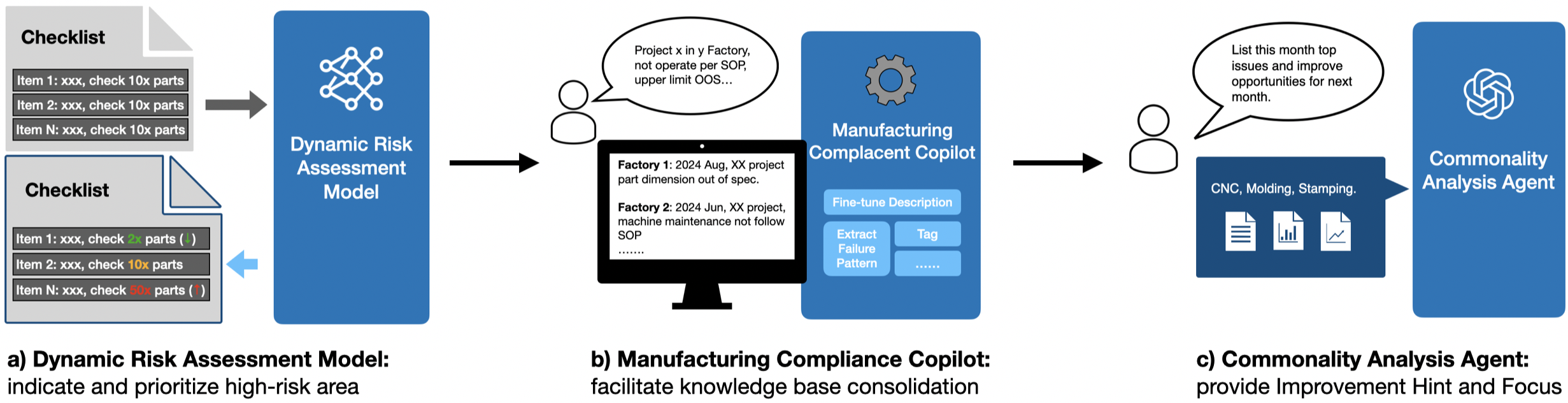}
    \caption{Smart Audit System Framework a) An initial phase utilizing natural language processing to optimize data collection through dynamic risk assessment; b) The integration of a compliance copilot system into existing manufacturing knowledge bases, employing a multi-task, instruction-guided analysis to convert raw data into actionable insights; c) Deployment of a Commonality Study Agent that enhances real-time accountability and process improvements using a variant consistency approach.}
    \label{fig:enter-label}
\end{figure*}

A quality audit in manufacturing is a systematic examination designed to ensure that production processes adhere to predefined quality standards and regulatory requirements. This involves a thorough evaluation of production methods, materials, and final products to detect any deviations from quality norms. The primary objectives of a quality audit are to uphold product consistency, enhance operational efficiency, and guarantee compliance with product specifications.

Traditional audits rely on manual efforts and expert knowledge, which introduces potential for errors, inefficiency, and inconsistency. These conventional methods struggle with maintaining uniformity and compliance across varied locations and procedures. Moreover, the data gathered during these audits are often unstructured and text-heavy, complicating any attempt to analyze them for patterns or anomalies, which in turn hampers the identification of trends or prevalent issues.

Emerging technologies, particularly LLMs, offer transformative potential for these traditional systems. LLMs excel in natural language understanding, information extraction, and text analysis, enabling more sophisticated and automated analysis of the extensive textual data typically collected during audits. By integrating LLMs, audits can transition from labor-intensive and error-prone processes to more streamlined, accurate, and consistent operations.

These smart audit systems aim to revolutionize the audit landscape by adopting more efficient and effective practices. By using LLMs, the system can reduce reliance on manual processes, improve accuracy, and enhance the consistency of audit outcomes, thus addressing the core challenges of traditional audit approaches.

This paper makes significant contributions in three areas:

\textbf{Dynamic Risk Assessment Model}: We design a model that streamlines audit focus, adaptively adjusts audit sample sizes, and pinpoints critical-quality items for each audit, optimizing resource allocation for detecting and mitigating high-risk issues.

\textbf{Manufacturing Compliance Copilot}: Our system facilitates the setup of a manufacturing knowledge base for data processing, retrieval, and evaluation, enabling seamless integration of new information with existing databases to foster self-evolution of the knowledge base.

\textbf{Commonality Analysis Agent}: We employ an autonomous agent capable of conducting real-time analysis tailored to customized requests, thus equipping engineers with the necessary insights to drive supplier improvements effectively.

\section{Related Works}

\subsection{Digitization in Quality Management Systems (QMS) and Supplier Management}
The inadequacies of traditional storage systems in monitoring Supplier Corrective Action Required (SCAR) instances have led to significant challenges in visibility and tracking within Supplier Quality Management (SQM). Scholars like Zulkifli, Abd., and Sangeethavani (2015) emphasize the need for enhanced data management strategies to bolster the efficacy of SQM processes. \cite{7359052} Furthermore, Mustroph \& Rinderle-Ma (2024) explore the role of artificial intelligence in enhancing risk management, data governance, and compliance processes within QMS, signifying a shift towards more resilient organizational frameworks. \cite{mustroph2024designqualitymanagementbased}

\begin{figure*}[ht]
    \centering
    \includegraphics[width=1\linewidth]{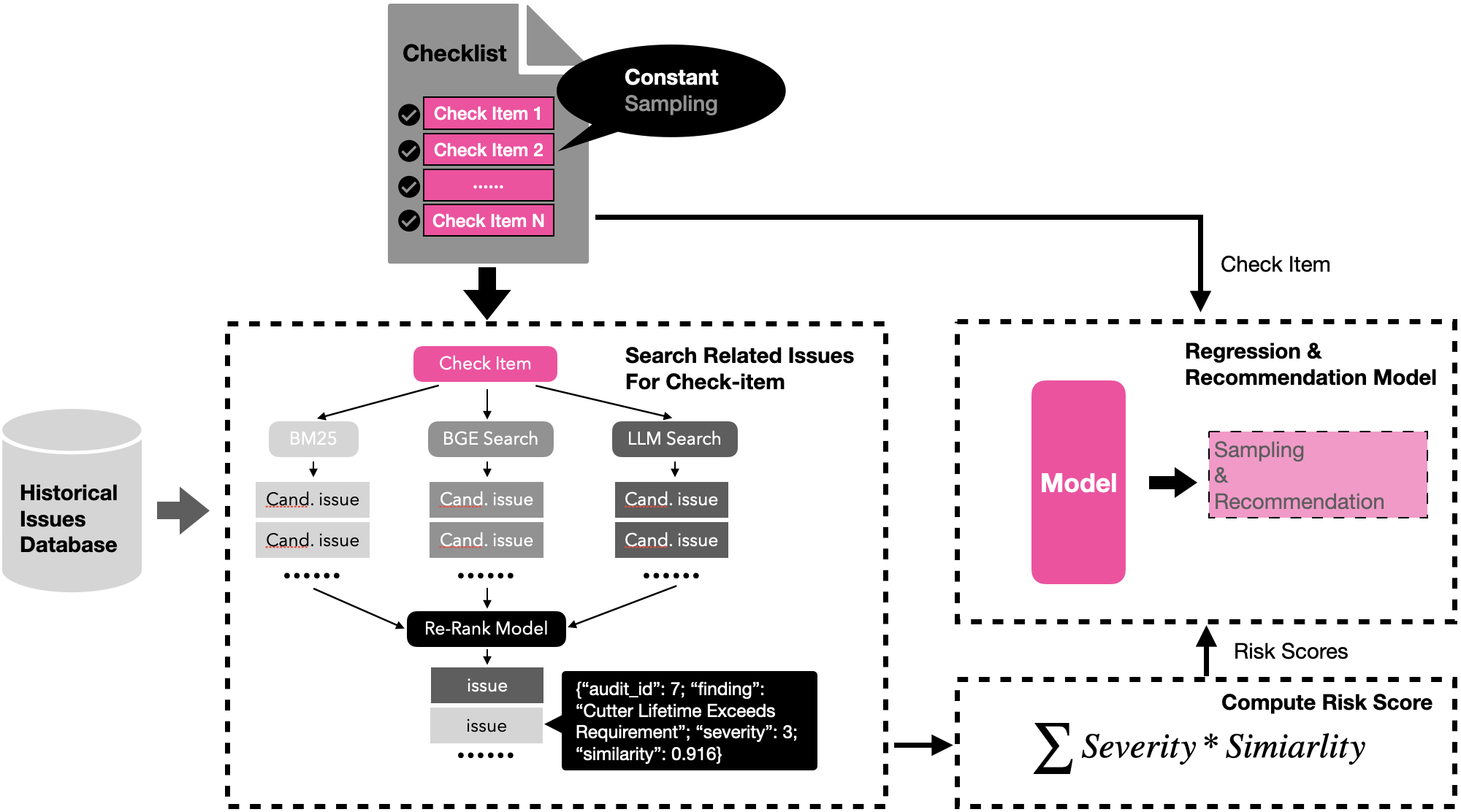}
    \caption{Dynamic Risk Assessment in Data Collection Pipeline}
    \label{fig:enter-label}
\end{figure*}

\subsection{Domain-Specified Retrieval-Augmented Generation (RAG)}
Retrieval-Augmented Generation (RAG). Existing research allows for the modular integration of knowledge bases into the generation systems of LLMs, enhancing their adaptability across various domains\cite{10.5555/3495724.3496517}. This modular approach not only enables the flexible attachment of diverse knowledge repositories, thus facilitating swift application in different fields \cite{gao2024modularragtransformingrag}, but also reduces the training demands associated with model fine-tuning. Importantly, it diminishes the incidence of 'hallucinations'—misleading or inaccurate outputs—commonly observed during the fine-tuning of LLMs on new knowledge \cite{gekhman2024doesfinetuningllmsnew}. 
Text Embeddings are instrumental in extracting latent features from original documents, significantly enhancing the accuracy of document retrieval and facilitating text categorization through clustering and classification algorithms \cite{Hen23}. In this study, the embedding models utilized include the BGE model \cite{luo2024bgelandmarkembeddingchunkingfree} and a customized variant of the all-MiniLM-L6-v2 model.\cite{Wang2020MiniLMDS} These models have been chosen specifically for their robust performance in handling linguistic data and enhancing semantic understanding within the framework of our analysis. 
Information Retrieval (IR). Information retrieval (IR) is a pivotal element of this project, enabling the tailored recall of original documents from a knowledge base to meet user-specific demands. \cite{Hambarde_2023} Prior research has underscored the efficacy of employing fused ranking systems for the retrieval and ordering of documents, which plays an essential role in the integrated search across multi-domain databases. \cite{10.1145/1571941.1572114} This approach allows the system to holistically evaluate various factors to output the most relevant documents. Within this framework, the project uses embedding semantic search based on cosine similarity. \cite{cakaloglu2019textembeddingsretrievallarge} alongside the BM25 ranking algorithm \cite{10.5555/188490.188561}, to enhance the precision of document retrieval.
Prompt Engineering. In the context of LLM Generation-centric systems, prompt engineering is identified as a pivotal component. Prior research has demonstrated that prompt engineering can significantly enhance task performance without the need for extensive retraining \cite{sahoo2024systematicsurveypromptengineering}.
In this study, the Chain-of-Density (CoD) approach is used to convert extensive texts from the knowledge base into a collection of concise summaries. \cite{adams2023sparsedensegpt4summarization} This strategy is specifically designed to rectify the challenges associated with the inaccurate matching of long and short texts during the text matching process.

\subsection{Re-Act Agent}
This study extends beyond conventional general prompt engineering by incorporating ReAct Prompting, which integrates methodologies such as Chain-of-Thought (CoT) and Function Calling. \cite{yao2022react} Notably, ReAct Prompting enhances CoT by infusing external knowledge, thereby facilitating more logical deliberation of documents within the knowledge base by LLMs. In the implemented system, function calling \cite{kim2024llmcompilerparallelfunction} is leveraged as a standalone mechanism to process data through dedicated code invocations. This method specifically facilitates the generation of detailed image analysis reports, employing isolated computational routines to enhance data handling efficiency.

\section{Methodologies}

\subsection{Overview}

Our methodology establishes a three-tiered system to revolutionize manufacturing audits using advanced technological frameworks, as depicted in Figure 1. 

Beginning with data collection phase, we implement a dynamic risk assessment model that employs natural language processing techniques to enhance the relevance and precision of audit data (a). This model utilizes an iterative analysis process, adapting to emerging data trends and historical insights to optimize data collection strategies.

Subsequently, we introduce a compliance copilot system that integrates seamlessly into existing manufacturing knowledge bases (b). This system employs a multi-task instruction-guided analysis pipeline to process, retrieve, and evaluate audit data, ensuring the effective transformation of raw data into actionable insights. This step involves sophisticated algorithms and LLM-enhanced tools to ensure data integrity and relevancy.

Finally, we deploy a Commonality Study Agent that operates autonomously to provide real-time accountability and process improvements (c). This agent leverages a variant consistency approach, using prompt-based instruction sets to generate tailored actions and reports, thus facilitating ongoing enhancements in audit processes and supplier performance.

Each component of our methodology is designed to synergistically enhance the efficiency and effectiveness of quality audits. Further details of each stage will be discussed in the subsequent sections.

\subsection{Dynamic Risk Assessment in Data Collection}

The dynamic risk assessment uses algorithms to differentiate high-risk and low-risk items in audit checklists, enabling engineers to allocate more resources to high-risk items and save time on low-risk ones.

By integrating semantic search, risk prioritization, and regression and recommendation models, the results could directly provide feedback to engineers through end device.

First, We begin by employing natural language processing (NLP) techniques to conduct a semantic search, linking historical audit issues with current check items to calculate the similarity between them.

Subsequent to the semantic linking, we compute Risk Priority Numbers (RPN) for each identified risk. Despite the occurrence (similarity) from semantic search, this computation also considers factors such as severity and detectability of risks. By prioritizing based on these criteria, we ensure that high-risk items can be flagged.

Finally, we use predictive model to direct our audit focus towards the highest-risk areas. This model leverages both regression and recommendation algorithms to analyze the prioritized data, enabling a targeted and strategic approach to data gathering within the audit process.

This optimized dynamic audit is for supporting the overall goals of facilitating more effective risk assessment and resource allocation in conducting compliance audit activities.

\subsection{Manufacturing Compliance Copilot in Knowledge Base Consolidation}

\begin{figure}[h]
    \centering
    \includegraphics[width=1\linewidth]{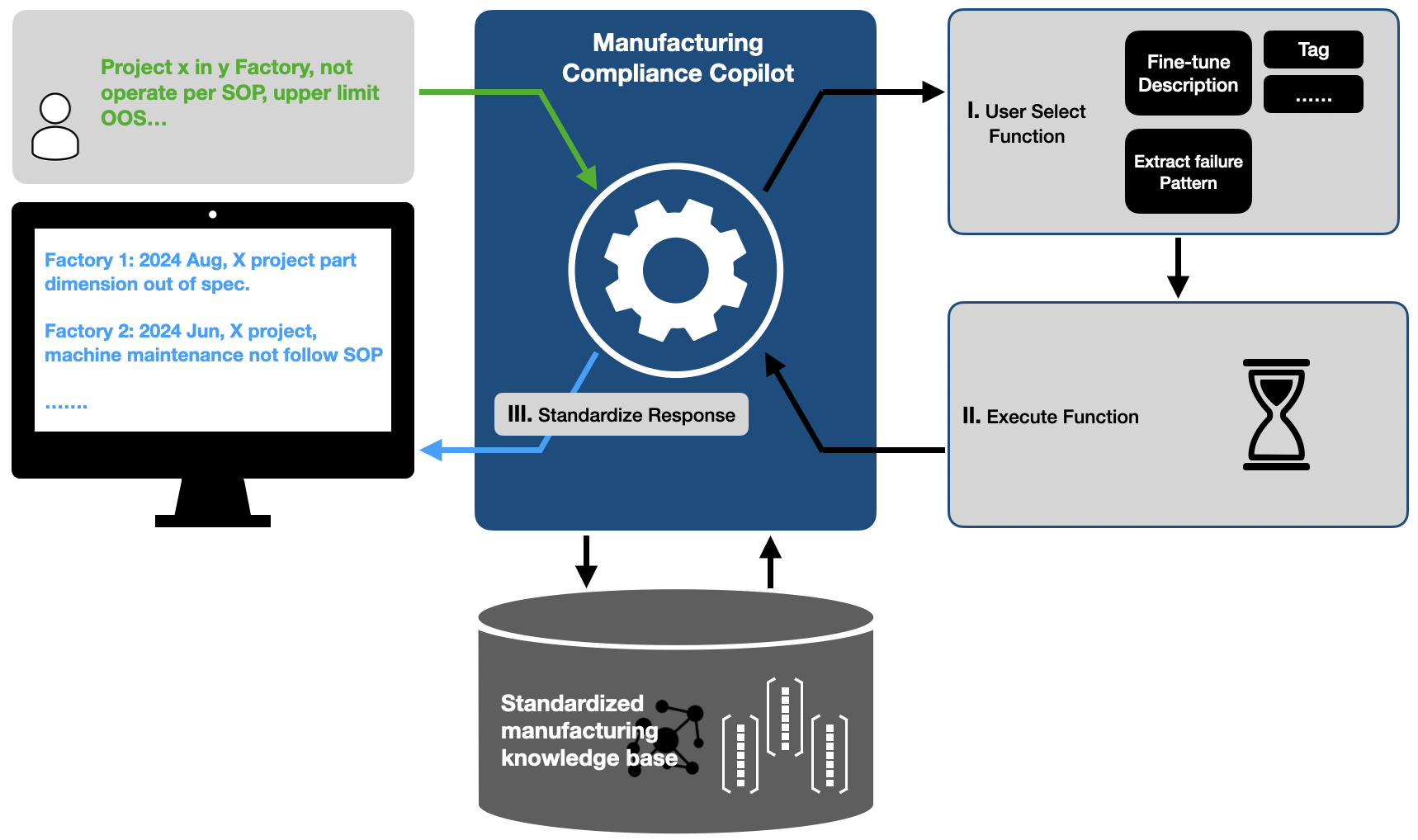}
    \caption{Manufacturing Compliance Copilot Structure}
    \label{fig:enter-label}
\end{figure}

The manufacturing compliance copilot aims to assist users in transforming raw data into a knowledge base that can be shared and utilized by others. This system-wise LLM-empowered copilot helps us optimize our database to create a more comprehensive and domain-specific manufacturing knowledge base.

This was achieved through an integrated approach that includes data processing, retrieval, and evaluation. A manufacturing compliance copilot streamlines the process, enhancing the manufacturing knowledge base to ensure seamless integration and self-evolution of data.

Data are initially processed using a language model that assists engineers by automating issue description tuning, failure pattern extraction, and tagging with predefined categories. This standardization ensures that all data entering the system is accurately scrutinized.

For data retrieval, we utilize a Retrieval-Augmented Generation (RAG) framework, which is enhanced with domain-specific embeddings to convert diverse data types into a uniform vector format. This setup supports hybrid retrieval methods that combine Semantic Search and BM25, supplemented by Reciprocal Rank Fusion (RRF) for effective reranking. A LLM, performs the final analyses and summarization. This enables engineers to review similar issues and root causes more thoroughly, offering an in-depth and comprehensive examination based on historical data retrieval.

Data quality evaluation is streamlined through DSpy, which quantifies subjective engineer opinions into objective comparable scores. This supports engineers in evaluating and monitoring the root cause analysis and corrective action quality across different suppliers with relatively equal standard.

This optimization seamlessly integrates into audit flow could improve the knowledge base quality that enabling all personal engineer experienced knowhow could turn into insights for everyone.

\subsection{Commonality Analysis Agent in Data Analysis}

\begin{figure}[h]
    \centering
    \includegraphics[width=1\linewidth]{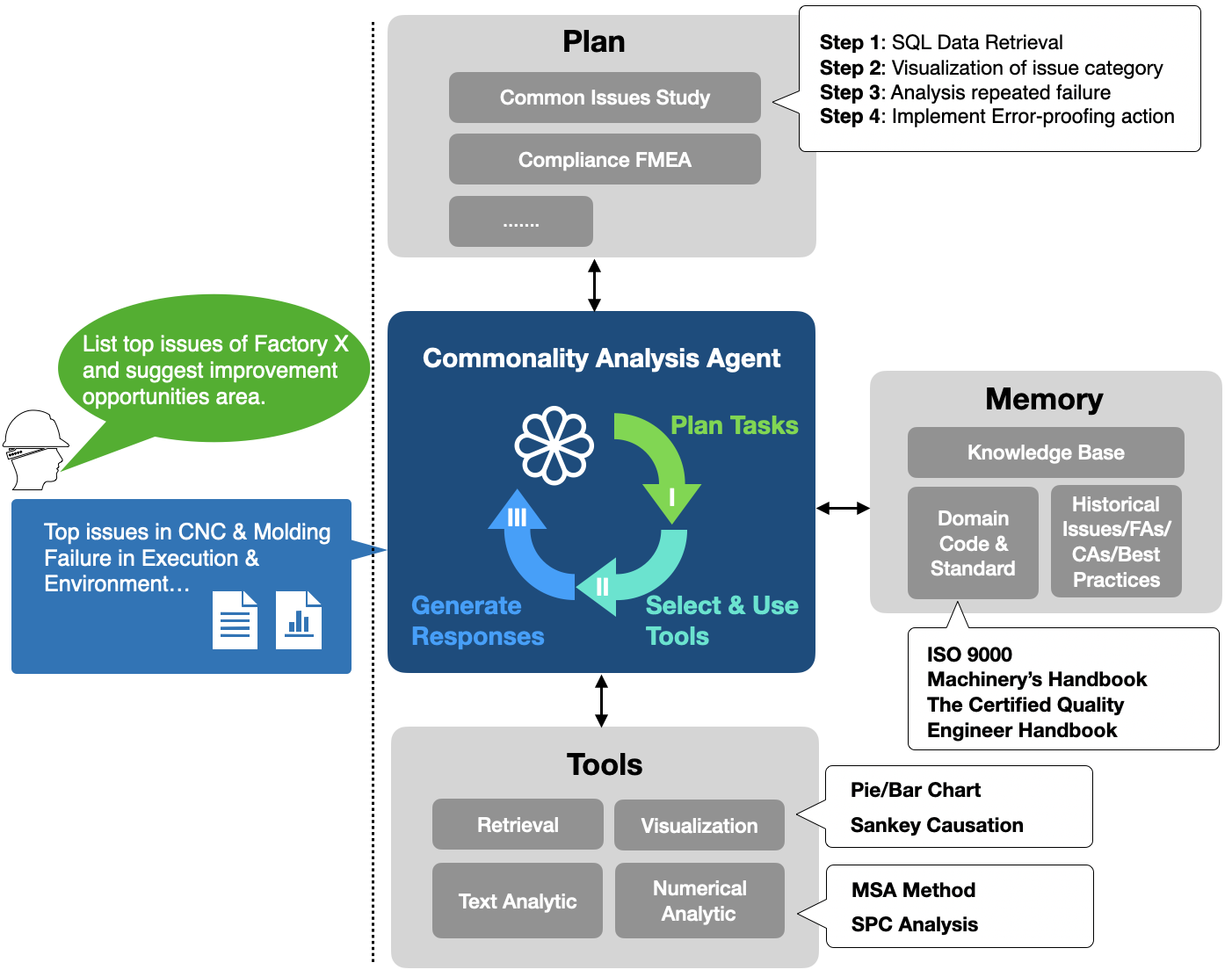}
    \caption{Commonality Analysis Agent Data Pipeline}
    \label{fig:enter-label}
\end{figure}

To enhance the capabilities of quality engineers in real-time analysis and supplier improvement, we implemented the Re-act framework agent, designed as an autonomous system that adapts to customized requests, providing targeted insights and recommendations required to drive supplier enhancements. This agent shall encompasses three key components: memory, plans, and tools.

\textbf{Memory}: The agent's memory includes industry standards and codes, proprietary company documentation, and a historical archive of failures and solutions. This dataset allows the agent to access relevant information swiftly, enhancing its decision-making capabilities.

\textbf{Plans}: The agent could autonomously generates and executes plans based on prompt-based instructions that guide its analytical processes. For example, it could employ the 5 Whys methodology for failure analysis and autonomously creates 8D reports outlining corrective actions with the help of predefined prompt.

\textbf{Tools}: The agent is equipped with data retrieval, numerical and text analysis, and visualization and reporting tools. These enable the agent to fetch data, perform analyses, and generate reports, supporting its analytical capabilities.

The operation of the Re-act agent begins with the user intention recognition module, which interprets the user's request and defines the specific task. The agent follows a structured, step-by-step plan to address the request. This automated process not only reduces the over-reliance on personnel knowledge and experience but also alleviates the labor-intensive aspects of reporting. By dynamically adapting its approach based on real-time human feedback, the agent ensures the provision of relevant and timely insights, easing the workload of engineers and streamlining the reporting process.

\section{Experiments}

\subsection{Setup}

For the experiment, we designed a test setup that evaluates the practical application of our proposed smart audit system in real-world settings. We leverage a LLM with 30-billion parameters to fit within the dynamic risk assessment model, manufacturing compliance copilot, and the commonality analysis agent.

To evaluate the efficacy of the smart audit system, we recruited 10 experienced quality auditors to interact with the system. These auditors, familiar with traditional auditing processes, were briefed on the system’s functionalities and asked to use it in parallel with their conventional auditing methods. This approach allowed us to directly assess the system’s enhancement of the audit process in terms of accuracy, time efficiency, and user satisfaction.

The following are the evaluation metrics

\textbf{Risk Prediction Accuracy}: Evaluates how well the system prioritizes audit tasks.

\textbf{Data Integrity Success Rate}: Assesses the accuracy with which the system categorized, tagged and retrieved data.

\textbf{User Satisfaction Score}: Collected through post-use questionnaires to gauge auditors' satisfaction with the system’s interface, ease of use, and overall utility, with a maximum of 100.

\textbf{Time Efficiency Improvement}: Measures the time saved in completing audit tasks using the system compared to traditional methods.

\subsection{Results}

The results from the real-world user testing are summarized in Table 1:

\begin{table}[h!]
\setlength{\tabcolsep}{5pt}
\centering
\caption{Evaluation Matrix of Smart Audit System}

\begin{tabular}{ll}
\hline
Metrics & Results \\
\hline
Risk Prediction Accuracy    & 90\% \\
Data Integrity Success Rate & 88\% \\
User Satisfaction Score     & 90   \\
Time Efficiency Improvement & 24\% \\
\hline
\end{tabular}
\end{table}

Our system demonstrated significant improvements in audit efficiency and accuracy. The auditors reported a 24 \% reduction in time spent on audits, which they attributed to the system's automated data processing and risk assessment capabilities. The high user satisfaction score reflects the system's ease of integration and operation within existing audit workflows.

\subsection{Use Cases Demonstrations}

\subsubsection{Smart audit app}

As shown in Figure 5, the dynamic audit algorithms would demonstrate and manifest in the end device, adjusting the sampling and highlighting the historical failure patterns.
\begin{figure}[h]
    \centering
    \includegraphics[width=1\linewidth]{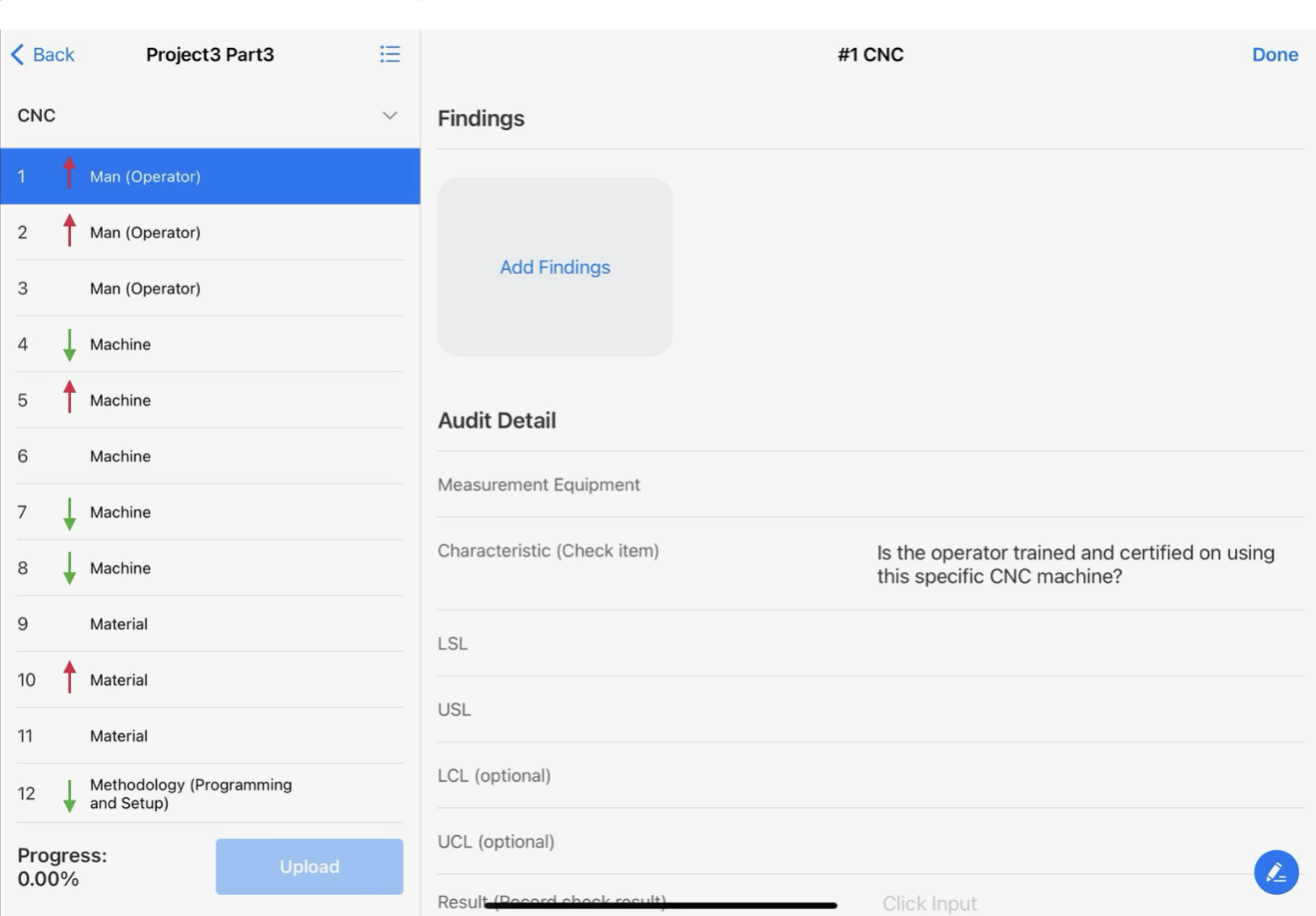}
    \caption{Smart Audit App: Facilitating Data Collection with an Embedded Dynamic Risk Assessment Model}
    \label{fig:enter-label}
\end{figure}

\subsubsection{Smart audit management web}

As illustrated in Figure 6, the audit record would be uploaded to the server for analysis by the compliance copilot, which includes description fine-tuning, categorization, tagging, and more.
\begin{figure}[h]
    \centering
    \includegraphics[width=1\linewidth]{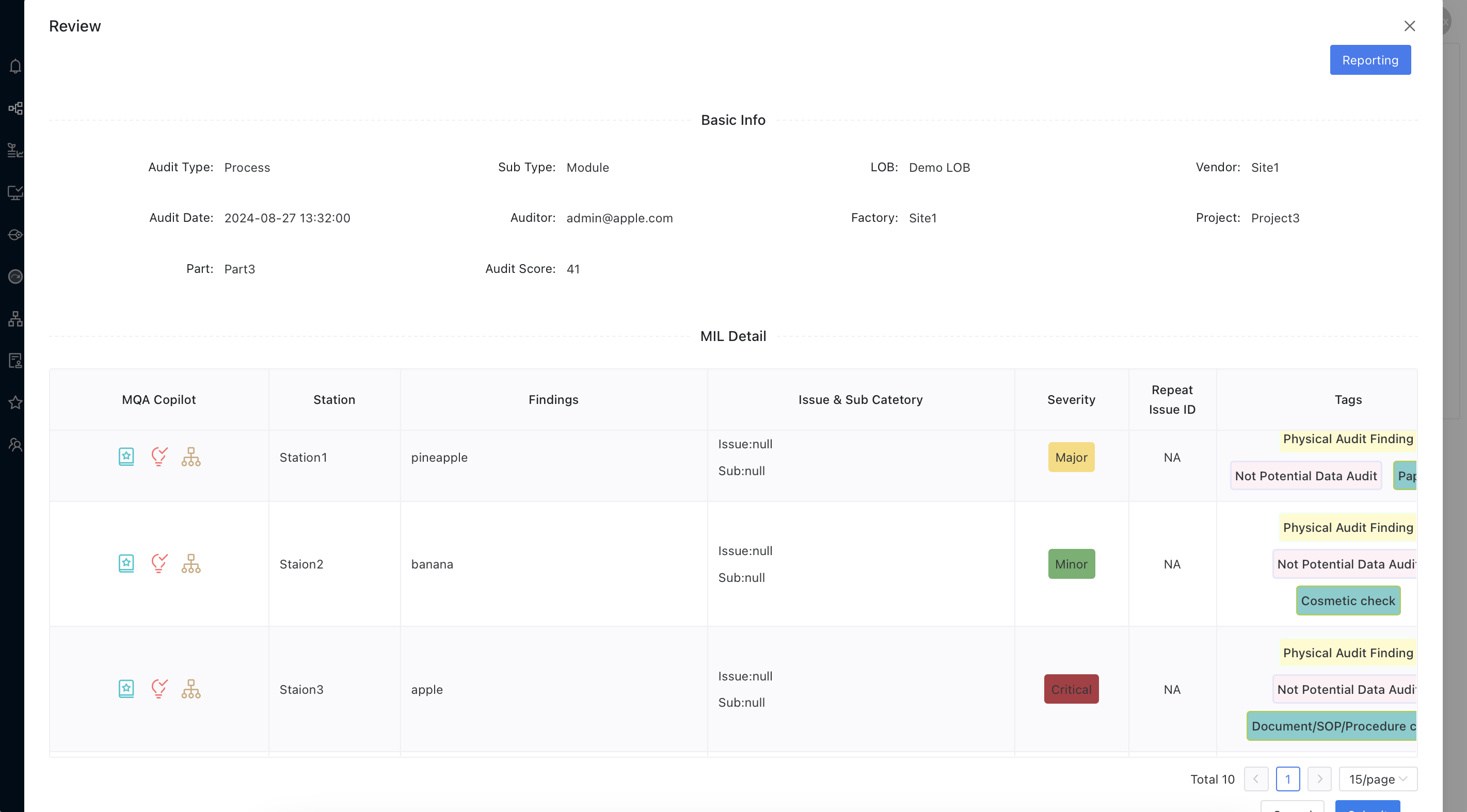}
    \caption{Smart Audit Management Web Platform with a System-Wise Embedded Copilot to Facilitate Knowledge-Based Consolidation and Commonality Analysis}
    \label{fig:enter-label}
\end{figure}

\subsection{Discussion}
The experiment confirmed the smart audit system's utility in enhancing the auditing process, though some areas for improvement were identified. The system occasionally struggled with highly complex audit scenarios, indicating a need for further refinement of its predictive algorithms to handle outlier cases more effectively. Future study will focus on enhancing the system’s adaptability and learning capabilities to improve handling of diverse audit conditions and further boost user satisfaction and efficiency.

\section{Conclusion}

In this paper, we addressed the complexities of traditional manufacturing quality audits by developing a smart audit system empowered by LLMs. 

To tackle the inefficiencies inherent in conventional auditing processes, we introduced a three-pronged approach: the dynamic risk assessment model, the manufacturing compliance copilot, and the Re-act framework agent. 

Initially, our system leverages NLP techniques to enhance data collection, ensuring that audit tasks are prioritized based on real-time risk assessments. Subsequently, the copilot integrates seamlessly with existing manufacturing databases, employing sophisticated data retrieval and analysis mechanisms to improve decision-making and ensure data integrity. Finally, the Re-act framework agent offers autonomous real-time analysis, adapting to specific auditing requirements to provide actionable insights and foster continuous improvements.

Experimental results from real-world applications demonstrate that the system significantly increases audit efficiency and accuracy, achieving high marks in data integrity success rate, risk prediction accuracy, and user satisfaction, while also reducing the time required for audit completion by 25 \%. These improvements highlight the system's potential to transform audit processes within the manufacturing sector.

\section{Limitations}

While our smart audit system significantly enhances the efficiency and accuracy of manufacturing audits, it currently exhibits limitations in handling highly irregular or outlier audit scenarios. Specifically, the system's performance can be less robust when encountering data patterns that significantly deviate from typical conditions, such as those found in unusual manufacturing defects or rare compliance issues. Furthermore, the system relies heavily on the quality and structure of the input data; discrepancies in data formatting or incomplete datasets can impede its effectiveness.

Future work will aim to enhance the system's ability to more effectively handle out-of-distribution data by developing algorithms that can detect and adapt to anomalies. We will also explore methods to improve system resilience, particularly in environments with high variability in audit conditions and requirements. 


\newpage

\bibliography{custom}

\end{document}